\documentclass[10pt,twocolumn,letterpaper]{article}

\usepackage{cvpr}              
\definecolor{cvprblue}{rgb}{0.21,0.49,0.74}
\usepackage[pagebackref,breaklinks,colorlinks,allcolors=cvprblue]{hyperref}

\def\paperID{19} 
\def\confName{CVPR}
\def\confYear{2026}

\title{Zero-Shot Object Re-Identification in Egocentric Kitchen Videos \\ via Multi-Stage SAM3 Feature Fusion}

\author{
Dmytro Klepachevskyi \quad
Alexander Wong \quad
Sirisha Rambhatla \quad
Yuhao Chen \\
University of Waterloo \\
Waterloo, Ontario, Canada \\
{\tt\small \{dklepachevskyi, alexander.wong, sirisha.rambhatla, yuhao.chen1\}@uwaterloo.ca}
}

\begin{document}
\maketitle
\begin{abstract}


Object re-identification (ReID) in egocentric kitchen videos is challenging due to rapid viewpoint changes, frequent occlusions, cluttered scenes, and large intra-class appearance variations. Objects may leave and re-enter the field of view, and the large diversity of instances with limited annotations makes supervised ReID difficult to scale, motivating zero-shot approaches. We study zero-shot object ReID on the EPIC-Kitchens benchmark, where the goal is to match active food and kitchen-tool instances across frames using only pre-trained visual features.
We first evaluate five state-of-the-art feature extractors, including Vision-Language Models (VLMs) - CLIP, DINOv2, DreamSim, I-JEPA, and SAM3 - and show that zero-shot methods fail, with the best baseline achieving only 45.3\% mAP. We then propose an Enhanced SAM3 ReID Pipeline, a zero-shot multi-stage method built around SAM3 segmentation as the core component. Stage 1 uses SAM3 to suppress background clutter. Stage 2 fuses embeddings from SAM3, DINOv2, and CLIP into a single L2-normalized descriptor. Stage 3 augments cosine similarity with mask-shape IoU for geometric consistency, and Stage 4 applies k-reciprocal re-ranking. The full pipeline improves performance by 7.5\% mAP to 52.8\%.

\end{abstract}    
\begin{figure}[t]
  \centering

   \includegraphics[width=\linewidth]{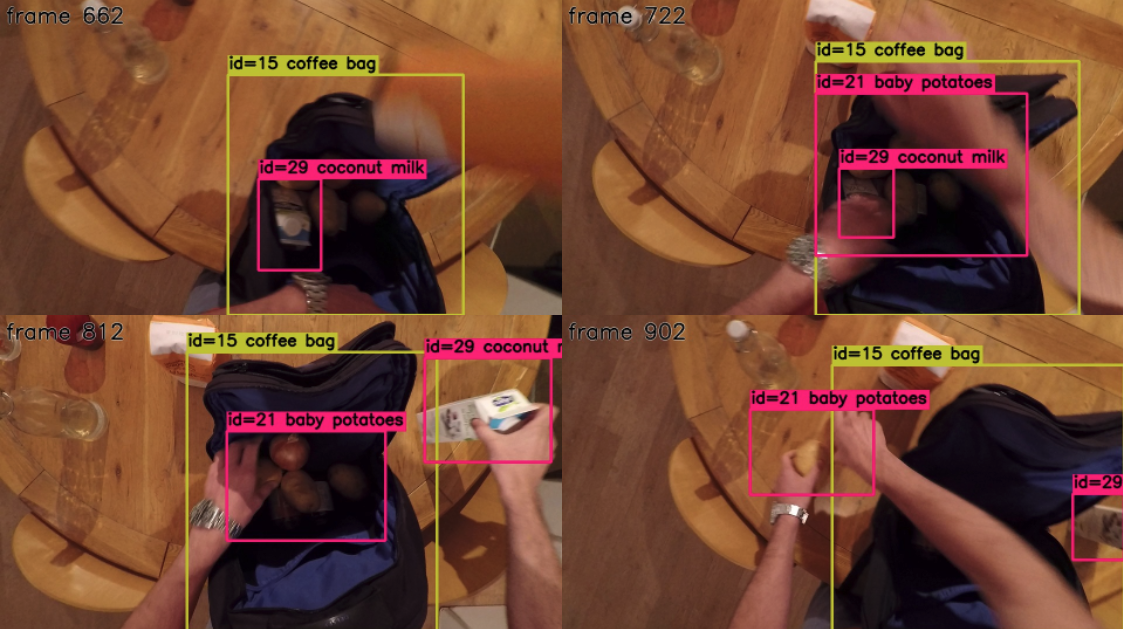}

   \caption{Example of a video sequence from EPIC-KITCHENS dataset with annotated objects. The ID of the same object is preserved across all the frames.}
   \label{fig:dataset}
\end{figure}

\section{Introduction}
\label{sec:intro}

Knowing \emph{which} specific food item a person is handling - and being able
to find it again later in the video - is a prerequisite for a broad class of
applications.
Automated dietary assessment systems must link the same apple or bowl of pasta
across multiple frames to estimate portions and log nutrition~\cite{Jia2019Egocentric,Thames2021Nutrition5k}.
Physically-grounded 3D food reconstruction, requires matching the same food object across
viewpoints before any volume estimation can occur.
Fine-grained cooking activity recognition~\cite{Rohrbach2012Cooking,Damen2022EPICKITCHENS100}
depends on knowing which ingredient is being cut, stirred, or plated at each
moment - a question that is fundamentally one of object identity rather than
category.
Yet despite this shared dependency, \emph{object re-identification} (ReID) in
kitchen video has received almost no dedicated study.

We address this gap.
Given a query crop of a kitchen object - a bowl, a spatula, a tomato -
the ReID task asks: can a system retrieve all other crops of the
\emph{same physical instance} across the video?
We study this in the zero-shot setting on EPIC-Kitchens~\cite{Damen2018EPICKITCHENS},
the largest egocentric kitchen benchmark, where ground-truth bounding-box
tracks define object identity but no identity-labelled training data is used.
 
This can play a crucial role in robotics applications, where a camera-based robot should be able to retrieve and find objects given a query. 

Person ReID \cite{person_reid_1, person_reid_2, person_reid_3, person_reid_4} and Vehicle ReID \cite{object_reid_1, object_reid_2} has been thoroughly studied,
but generalizable object ReID in kitchen settings introduces a distinct set of challenges.
First, objects are \emph{semantically diverse}: a ``cutting board'' or a
``handful of pasta'' has no canonical orientation or colour, unlike a pedestrian.
Second, objects experience \emph{extreme partial occlusion} as hands manipulate
them, and backgrounds are cluttered with other food and surfaces.
Third, and most importantly for this work, while ground-truth bounding-box
tracks are available for evaluation, no identity-labelled kitchen-object
data is used for training - all models are used zero-shot.

Recent advances in vision-language pre-training and self-supervised learning
have produced powerful state-of-the-art visual encoders.
CLIP~\cite{Radford2021CLIP} aligns visual and linguistic semantics,
DINOv2~\cite{Oquab2024DINOv2} learns instance-discriminative patch features,
DreamSim~\cite{Fu2023DreamSim} targets human perceptual similarity,
I-JEPA~\cite{Assran2023IJEPA} predicts latent representations,
and SAM3~\cite{SAM3_2025} extends Segment Anything to video with a powerful
vision-language backbone.
Each encoder captures a different facet of visual appearance, motivating
ensemble strategies.

\begin{figure*}[t]
  \centering

   \includegraphics[width=\textwidth]{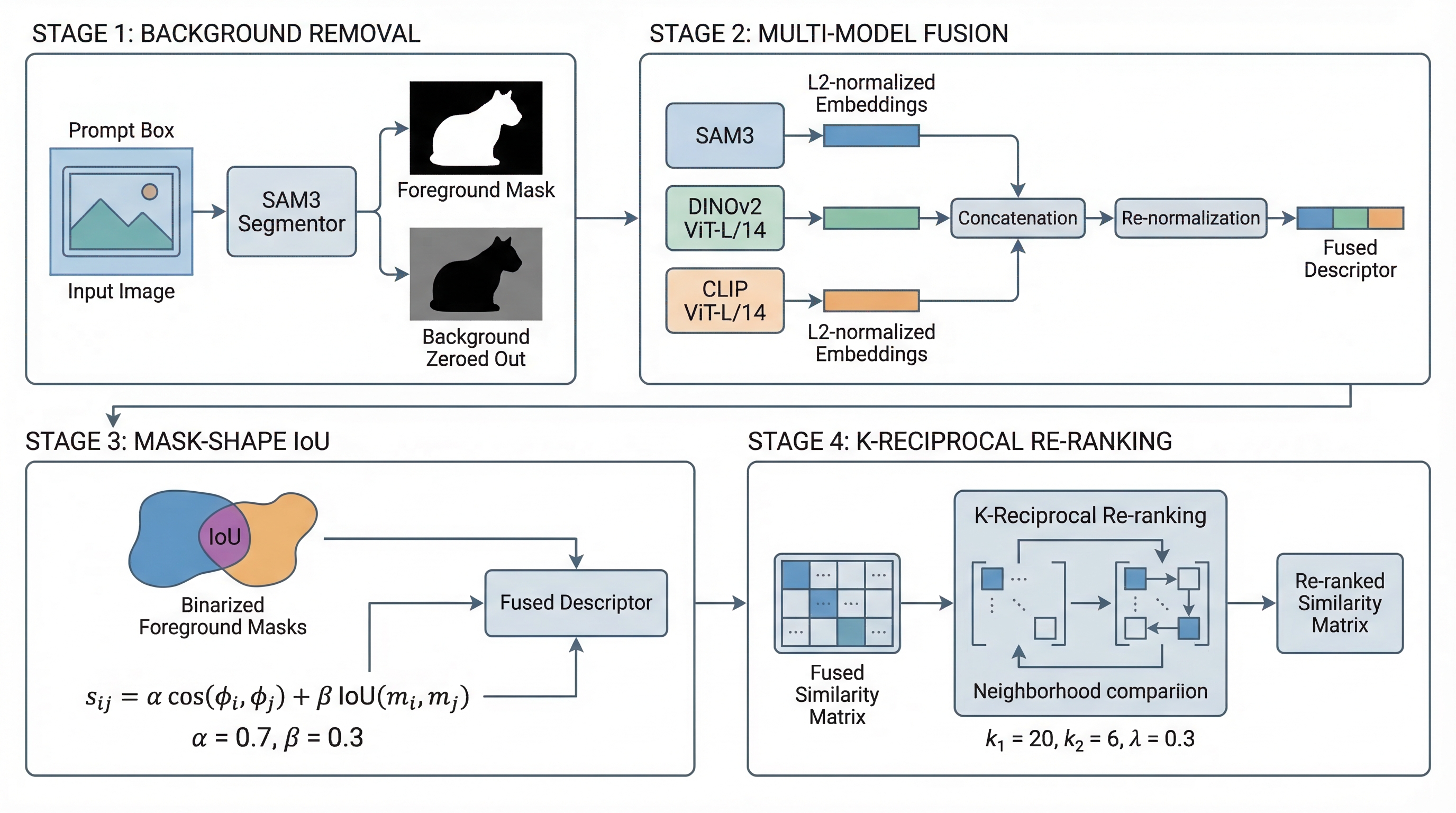}

   \caption{Our 4-staged enhanced SAM3 method with feature fusion. The first stage incorporates  background removal to get a foreground mask. The second stage uses a multi-model fusion of three embeddings (SAM3, DINOv2, CLIP). The Stage 3 uses cosine similarity with an IoU term. The Stage 4 applies K-reciprocal re-ranking to the fused similarity matrix.}
   \label{fig:method}
\end{figure*}

In this paper we make three contributions:
\begin{enumerate}[noitemsep,topsep=2pt]
  \item We establish the first systematic zero-shot ReID benchmark on
        EPIC-Kitchens (Figure \ref{fig:dataset}), evaluating six state-of-the-art encoders under a
        unified protocol (\S\ref{sec:experiments}).
  \item We design and ablate a four-stage Enhanced SAM3 pipeline (Figure \ref{fig:method})
        (background removal, multi-model fusion, mask-IoU reweighting,
        k-reciprocal re-ranking), which achieves mAP of 0.528, which leads to an improvement of 7.5\%.
  \item Guided by the ablation, we also propose a lightweight
        \emph{Multi-Model Fusion} baseline that achieves
        mAP of 0.460 - a 0.7\% relative improvement over the
        best other single encoder baseline - using only L2-normalised feature concatenation
        on unmodified crops (\S\ref{sec:experiments}).
\end{enumerate}

Our findings have a clear practical message: a complex segmentation-based preprocessing pipeline shows significant improvements in zero-shot egocentric ReID, and diverse feature ensembles on natural crops set up strong baseline results.




\section{Methodology}
\label{sec:methodology}

\subsection{Method Motivation}
\label{sec:motivation}

Standard multi-object trackers handle short-term occlusion well but break down across
long temporal gaps, exactly the scenario that arises when a pot is moved off-screen
or an ingredient is set aside and retrieved later. Closing this gap requires a
retrieval-based approach: given a crop of an object at one moment in time, find all
other crops of the same object in a large gallery, regardless of when or how they appear.

Our work addresses this gap in a zero-shot setting, without any fine-tuning on
kitchen data. This choice is deliberate: annotated identity tracks in egocentric
kitchen video~\cite{Damen2022EPICKITCHENS100} are scarce and expensive to produce, so a practical system must generalize from large-scale pre-trained representations. We systematically evaluate which
state-of-the-art visual encoders transfer best to this domain, and we design a
post-processing pipeline that squeezes additional performance from these frozen
features through background suppression, complementary fusion, and graph-based
re-ranking.

\subsection{Problem Formulation}
Given a video with $N$ ground-truth bounding-box crops
$\mathcal{X} = \{x_i\}_{i=1}^{N}$, each assigned a track identity
$y_i \in \{1,\ldots,K\}$, we split $\mathcal{X}$ into a gallery set
$\mathcal{G}$ (75\%) and a query set $\mathcal{Q}$ (25\%) by stratified
random sampling.
For each query $q \in \mathcal{Q}$ we rank all gallery items
$g \in \mathcal{G}$ by similarity and evaluate with mAP and Top-$k$ metrics.
All methods are \emph{zero-shot}: no identity labels are used at any stage.

\subsection{Single-Model Baselines}
\label{sec:baselines}

We extract a fixed-size embedding $\phi(x) \in \mathbb{R}^d$ from each crop
using six encoders, all used off-the-shelf with no fine-tuning.
The encoders span a broad design space, from language-supervised to
purely self-supervised models, and from general-purpose to task-specific backbones.

We begin with \textbf{CLIP ViT-B/32}~\cite{Radford2021CLIP}, which encodes each
crop with the visual branch of a vision-language model pre-trained on 400M
image-text pairs, producing a CLS-token embedding of dimension $d\!=\!512$.
As a purely self-supervised alternative we evaluate \textbf{DINOv2 ViT-B/14}~\cite{Oquab2024DINOv2},
trained with knowledge distillation on a curated 142M-image corpus ($d\!=\!768$),
and its successor \textbf{DINOv3 ViT-B/16}, which extends the same recipe to a
larger and more diverse pre-training set.
A qualitatively different signal comes from \textbf{DreamSim}~\cite{Fu2023DreamSim},
a perceptual similarity model that combines CLIP, DINO, and OpenCLIP embeddings
via a learned MLP head trained on human triplet judgements, making it sensitive
to the kind of holistic appearance differences that matter to human observers.
We also include \textbf{I-JEPA ViT-H}~\cite{Assran2023IJEPA}, which learns by
predicting latent patch representations from masked context regions without
any pixel-level reconstruction; features are obtained by mean-pooling patch
tokens ($d\!=\!1280$). Finally, \textbf{SAM3 ViT-H}~\cite{SAM3_2025} is the
vision backbone of the Segment Anything 3 model, pre-trained on
large-scale video data; we extract vision features and apply global
average pooling ($d\!=\!256$). For all encoders, gallery items are ranked by
cosine similarity to the query embedding.

\subsection{Enhanced SAM3 Pipeline}
\label{sec:enhanced}

Error analysis of the single-model baselines reveals three recurring failure reasons: cluttered backgrounds that make embeddings noisy with irrelevant texture, the limited coverage of any individual pre-trained model, and cosine similarity scores that ignore the geometric shape of the object. No single encoder addresses all three simultaneously, which motivates a pipeline that tackles each failure reason. Building on these observations, we propose a four-stage pipeline that combines background suppression, multi-model feature fusion, geometry-aware scoring, and graph-based re-ranking.
Each stage targets a distinct source of error in zero-shot kitchen ReID, and together they form a modular system whose components can be used independently, as confirmed by our ablation study (Table~\ref{tab:ablation}).

The first stage addresses the fact that kitchen crops often contain
cluttered backgrounds — counter-tops, appliances, and hands — that
introduce spurious texture signals into the embedding.
For each crop, we invoke the SAM3~\cite{SAM3_2025} segmentor with a
prompt box covering the full crop extent (normalised to $[0,1]$) to obtain
a binary foreground mask $m_i \in \{0,1\}^{H \times W}$.
The background pixels are zeroed out before any feature is extracted,
so the encoder focuses exclusively on the object of interest.

In the second stage we compensate for the limited coverage of any
single pre-trained model by fusing three complementary encoders.
L2-normalised embeddings from SAM3 ViT-H, DINOv2 ViT-L/14, and
CLIP ViT-L/14 are concatenated along the feature dimension and
re-normalised to unit length, yielding a single fused descriptor:
\begin{equation}
  \Phi_i = \ell_2\!\left([\,\ell_2(\phi_i^{\text{SAM3}}) \;\|\; \ell_2(\phi_i^{\text{DINOv2}}) \;\|\; \ell_2(\phi_i^{\text{CLIP}})\,]\right).
  \label{eq:fusion}
\end{equation}
SAM3 provides instance-level spatial detail, DINOv2 contributes
robust semantic structure, and CLIP supplies open-vocabulary
object-class information, making the three encoders largely complementary.

The third stage enriches the pairwise similarity score with a
geometric signal derived from the foreground masks produced in Stage~1,
using the fused descriptor $\Phi_i$ from Eq.~\ref{eq:fusion}.
Pure cosine similarity treats two crops as equally similar regardless
of whether their object silhouettes agree.
We replace it with a weighted combination:
\begin{equation}
  s_{ij} = \alpha\,\cos(\Phi_i,\Phi_j) + \beta\,\text{IoU}(m_i, m_j),
  \label{eq:similarity}
\end{equation}
where the IoU term rewards pairs whose foreground shapes overlap well
and penalises matches that are visually similar in texture but geometrically
inconsistent. We set $\alpha\!=\!0.7$ and $\beta\!=\!0.3$ throughout all experiments.

The fourth and final stage applies k-reciprocal re-ranking~\cite{Zhong2017Rerank}
to the full pairwise similarity matrix defined by Eq.~\ref{eq:similarity}.
Re-ranking exploits the mutual-neighbour structure of the gallery:
two items are up-weighted if they appear in each other's top-$k$ neighbour
lists, which is a strong indicator of a true match.
We use the hyperparameters $k_1\!=\!20$, $k_2\!=\!6$, and $\lambda\!=\!0.3$,
following standard practice in person ReID literature.






\subsection{Multi-Model Fusion on Natural Crops}
\label{sec:fusion}

Motivated by the ablation, we propose another simple pipeline for results improvement:
\emph{fuse complementary encoders directly on the original, unmodified crop}.
Given feature parts $\{\phi_m(x)\}_{m=1}^{M}$ from $M$ encoders,
we compute:
\begin{equation}
  \Phi(x) = \ell_2\!\left(
    \left[\,\ell_2(\phi_1(x)) \;\|\; \cdots \;\|\; \ell_2(\phi_M(x))\,\right]
  \right),
\end{equation}
where $\|$ denotes concatenation and $\ell_2$ denotes row-wise L2 normalisation.
We evaluate all subsets of \{DINOv2, DreamSim, CLIP~ViT-L/14\}
and additionally test \emph{Average Query Expansion}
(AQE,~\cite{Chum2007AQE}) with $k\!=\!10$:
each query is replaced by the mean of itself and its top-$k$ gallery neighbours
before final ranking.

\section{Experiments}
\label{sec:experiments}

In this section, we provide both qualitative and quantitative results of our experiments. Qualitatively, we compare how different baseline models track an object ID in a video sequence. We provide qualitative results for 3 best baseline models (CLIP, DINOv2, SAM3) and for two proposed methods by our study --- Enhanced SAM3 pipeline and a Multi-Model Fusion method.

\subsection{Dataset}

We evaluate our pipeline on 10 mostly overcrowded with objects video sequences from the EPIC-KITCHENS\cite{Damen2018EPICKITCHENS} dataset (Figure \ref{fig:dataset}). 
Ground-truth bounding-box tracks from the MOT-format annotations define object identities.
Crops smaller than 32~px in either dimension are discarded.
A 5\% padding is applied around each bounding box before cropping.
The 75\%/25\% gallery/query split is fixed for all experiments.

\subsection{Metrics}

For each query crop, all gallery items are ranked by cosine similarity to the query embedding. We report two complementary metrics:

Mean Average Precision (mAP) measures the quality of the full ranking. For a query $q$ with $R$ relevant items in the gallery, Average Precision is defined as:
\begin{equation}
  \mathrm{AP}(q) = \frac{1}{R}\sum_{k=1}^{N} P(k)\cdot \mathrm{rel}(k),
  \label{eq:ap}
\end{equation}
where $N$ is the gallery size, $P(k)$ is the precision at cut-off $k$, and $\mathrm{rel}(k)\in\{0,1\}$ indicates whether the item ranked at position $k$ is a true match. mAP averages AP over all $Q$ queries:
\begin{equation}
  \mathrm{mAP} = \frac{1}{Q}\sum_{q=1}^{Q}\mathrm{AP}(q).
  \label{eq:map}
\end{equation}
mAP is the area under the Precision-Recall curve, rewarding systems that place all true matches at the top of the ranked list. Unlike Top-$K$, mAP penalises false positives anywhere in the ranking, making it the primary metric for retrieval evaluation.

Cumulative Matching Characteristic (CMC) at Top-1, Top-3, and Top-5 measures recall: the fraction of queries for which at least one true match appears within the top $K$ retrieved results. 

The two metrics capture different failure modes: a system can achieve high Top-1 by consistently finding one easy match, while mAP exposes whether it retrieves \emph{all} instances of an identity. We use both to provide a complete picture of retrieval performance.

\subsection{Single-Model Baseline Results}

\Cref{tab:baselines} summarises all single-model baselines on 10 mostly overcrowded sequences.
DreamSim achieves the best mAP (0.453) and Top-1 (85.2\%).
I-JEPA attains the third-highest Top-1 (81.7\%) despite weaker mAP,
suggesting it is good at nearest-neighbour retrieval but less precise
at ranking multiple positives. SAM3, having a powerful backbone, ranks second after DreamSim;
its features which are optimised for segmentation work fine for instance discrimination as well.
DINOv3, evaluated on ten sequences, performs poorly (mAP of 0.110),
suggesting that its larger pre-training distribution does not transfer well
to close-range kitchen objects without fine-tuning.

\begin{table}[h]
\centering
\caption{Single-model zero-shot ReID in comparison to our Enhanced SAM3 method on EPIC-Kitchens dataset.}
\label{tab:baselines}
\setlength{\tabcolsep}{5pt}
\begin{tabular}{lcccc}
\toprule
Model & mAP & Top-1 & Top-3 & Top-5 \\
\midrule
CLIP ViT-B/32      & 0.355 & 0.773 & 0.889 & 0.926 \\
DINOv2 ViT-B/14    & 0.416 & 0.774 & 0.903 & 0.939 \\
DINOv3 ViT-B/16    & 0.110 & 0.742 & 0.891 & 0.929 \\
I-JEPA ViT-H       & 0.321 & 0.817 & 0.913 & 0.942 \\
SAM3 ViT-H         & 0.451 & 0.872 & \textbf{0.946} & \textbf{0.955} \\
DreamSim    & 0.453 & 0.852 & 0.942 & 0.954 \\
Ours Enhanced SAM3 & \textbf{0.528} & \textbf{0.893} & 0.907 & \textbf{0.955} \\
\bottomrule
\end{tabular}
\end{table}

\begin{figure*}[t]
  \centering

   \includegraphics[width=0.65\textwidth]{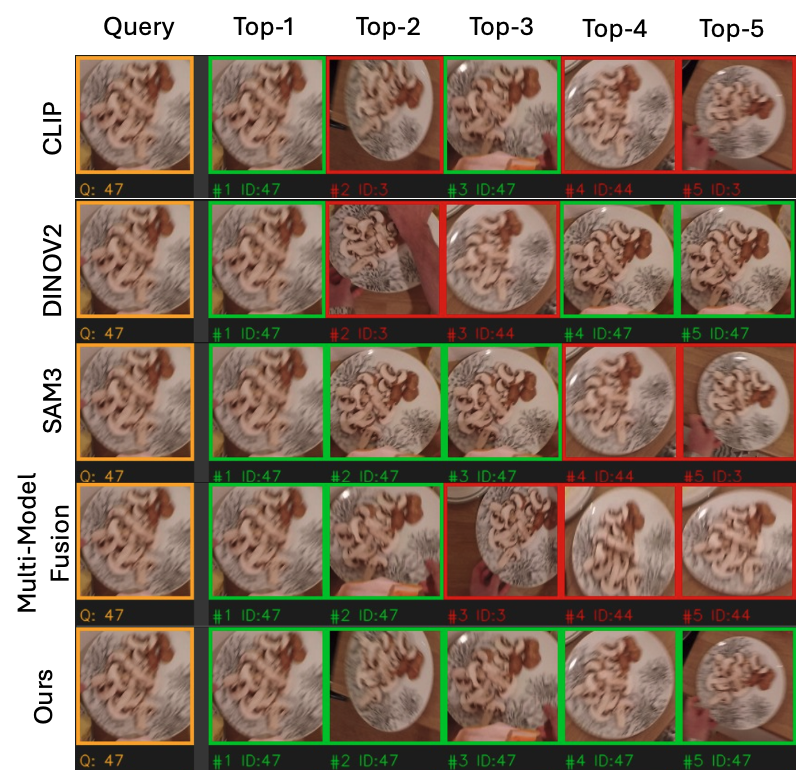}

   \caption{Qualitative results. We evaluate three baseline methods (CLIP, DINOv2, SAM3) along with our methods --- Multi-Model Fusion and Enhanced SAM3    Fusion method. The evaluation is presented given the same query object and Top-5 predictions. The analysis shows that 4 methods make at least 2 mistakes in the object ID assignment, while our SAM3 enhanced pipeline make all 5 correct predictions.}
   \label{fig:qualitative}
\end{figure*}

\subsection{Enhanced SAM3 pipeline}

The full pipeline (all stages enabled) achieves mAP of 0.528 (Table \ref{tab:baselines}) and Top-1 of 0.893 - which gives an improvement of 7.5\% of mAP metric, and 2.1\% of Top-1 metric over the best single-model baseline (DreamSim). Qualitative analysis (Figure \ref{fig:qualitative}) reveals that for given queries, the developed enhanced SAM3 Feature Fusion method significantly improves the results, where it manages to achieve all 5 correct retrievals, whereas the other methods fail at least 2 times.

\subsection{Ablation Study}

\Cref{tab:ablation} shows the full ablation of the four-stage Enhanced SAM3
pipeline (shown in Figure \ref{fig:method}). We quantitatively confirm that removal of the feature fusion part hurts the most (5.5\% drop of mAP), whereas removal of every single component hurts the performance, which proves that all 4 stages overall improve the performance.

\begin{table}[h]
\centering
\caption{Enhanced SAM3 pipeline ablation on EPIC-Kitchens dataset.
Each row disables one stage; $\checkmark$ = enabled.}
\label{tab:ablation}
\setlength{\tabcolsep}{4pt}
\begin{tabular}{ccccccc}
\toprule
BG & Fusion & MaskIoU & Rerank & mAP & Top-1 & Top-3 \\
\midrule
\checkmark & \checkmark & \checkmark & \checkmark & \textbf{0.528} & \textbf{0.893} & \textbf{0.907} \\
           & \checkmark & \checkmark & \checkmark & 0.501 & 0.842 & 0.863 \\
\checkmark &            & \checkmark & \checkmark & 0.473 & 0.827 & 0.849 \\
\checkmark & \checkmark &            & \checkmark & 0.511 & 0.884 & 0.897 \\
\checkmark & \checkmark & \checkmark &            & 0.498 & 0.835 & 0.862 \\
\bottomrule
\end{tabular}
\end{table}

\subsection{Multi-Model Fusion Results}

\Cref{tab:fusion} reports fusion experiments on ten sequences from EPIC-Kitchens dataset.
All runs use natural, unmodified crops.
DINOv2$+$DreamSim already outperforms every single baseline encoder (mAP of 0.455).
Adding CLIP provides a marginal further gain in mAP (0.458) with no significant Top-1 change.
Average Query Expansion (AQE, $k\!=\!10$) improves mAP slightly
(0.460 with 3-way fusion) but reduces Top-1
(0.801 vs.\ 0.825), as expanded queries become less discriminative in the small gallery setting.
The best balanced operating point is \textbf{DINOv2$+$DreamSim$+$CLIP}
without AQE: mAP 0.458, Top-1 82.5\%.

\begin{table}[h]
\centering
\caption{Multi-model fusion on ten sequences of EPIC-Kitchens dataset. All encoders run on original crops.}
\label{tab:fusion}
\setlength{\tabcolsep}{5pt}
\begin{tabular}{lcccc}
\toprule
Models & QE & mAP & Top-1 & Top-3 \\
\midrule
DreamSim                      &     & 0.433 & 0.822 & 0.932 \\
DINOv2$+$DreamSim             &     & 0.455 & 0.820 & 0.934 \\
DINOv2$+$DreamSim$+$CLIP      &     & 0.458 & \textbf{0.825} & \textbf{0.936} \\
DINOv2$+$DreamSim             & 10  & 0.447 & 0.789 & 0.879 \\
DINOv2$+$DreamSim$+$CLIP      & 10  & \textbf{0.460} & 0.801 & 0.877 \\
\bottomrule
\end{tabular}
\end{table}



\subsection{Conclusion}

We presented a study of zero-shot object re-identification in egocentric kitchen videos on EPIC-Kitchens. Across six state-of-the-art encoders, DreamSim and SAM3 emerge as the strongest single-model baselines (mAP $\approx$0.45), while encoders with larger pre-training corpora (DINOv3) do not necessarily transfer better to close-range kitchen objects.
Our four-stage Enhanced SAM3 pipeline - background removal, multi-model embedding fusion, mask-IoU reweighting, and k-reciprocal re-ranking - achieves mAP 0.528, a relative gain of 7.5\% over the best single encoder.
A lightweight Multi-Model Fusion baseline (DINOv2$+$DreamSim$+$CLIP) reaches mAP 0.458, which runs 5x times faster in wall-clock time per each query than the SAM3 enhanced method.
Ablations confirm that each stage contributes independently.
These results establish strong zero-shot baselines for kitchen-object ReID and highlight that background removal, complementary feature ensembling, and geometric re-ranking are beneficial stages for future work in egocentric instance retrieval.

{
    \small
    \bibliographystyle{ieeenat_fullname}
    \bibliography{main}
}


\end{document}